\PassOptionsToPackage{table}{xcolor}

\documentclass[11pt,letterpaper,logo]{thuair}
\newcommand{\arxivversion}{}

\graphicspath{{./}}

\usepackage[numbers,sort&compress]{natbib}

\usepackage[section]{placeins}


\usepackage{xspace}
\usepackage{graphicx}
\usepackage{subcaption}
\usepackage{multirow}
\usepackage{makecell}
\usepackage{amsmath}
\usepackage{amssymb}
\usepackage{enumitem}

\usepackage[ruled,linesnumbered,noend]{algorithm2e}
\usepackage{hyperref}
\usepackage{listings}
\newcommand{\nickname}{ActProbe\xspace}

\title{\nickname: Action-Space Probe for Early Failure Detection of Generative Robot Policies}

\makeatletter
\def\@fnsymbol#1{\ensuremath{\ifcase#1\or *\or \dagger\or \ddagger\or
    \mathsection\or \mathparagraph\or \|\or **\or \dagger\dagger
\or \ddagger\ddagger \else\@ctrerr\fi}}
\makeatother

\author{%
  Bingjia Huang$^{1,2*\dagger}$ \quad
  Xiangyu Li$^{1*}$ \quad
  Xiang Wang$^{1}$ \quad
  Liang Mi$^{3\dagger}$ \quad
  Zixu Hao$^{1}$ \\
  Weijun Wang$^{1}$ \quad
  Hao Wu$^{3}$ \quad
  Kun Li$^{1}$ \quad
  Yunxin Liu$^{1}$ \quad
  Ting Cao$^{1\ddagger}$ \\
  \medskip
  {
    $^{1}$Institute for AI Industry Research (AIR), Tsinghua University \\
    $^{2}$University of Electronic Science and Technology of China \quad
    $^{3}$Nanjing University \\
  }
  $^{*}$Equal contribution. \quad
  $^{\dagger}$Work done during internship at AIR, Tsinghua University. \\
  $^{\ddagger}$Corresponding author: Ting Cao (\href{mailto:tingcao@mail.tsinghua.edu.cn}{\textcolor{black}{tingcao@mail.tsinghua.edu.cn}}). \\
  \medskip
  \textbf{Project Page:} \url{https://air-embodied-brain.github.io/actprobe} \\
  \textbf{Source Code:} \url{https://github.com/air-embodied-brain/actprobe}
  \vspace{-12pt}
}

\begin{document}

% Redefine thuair's first-page footer to include the *, †, ‡ explanations BELOW
% the thuair gray footer rule (instead of as LaTeX footnotes above it).
% \fancypagestyle{firststyle}{%
%   \fancyhead[L]{\includegraphics[width=140pt]{thuair_title.png}}
%   \fancyhead[R]{\footerfont\itshape\monthyeardate\today}
%   \fancyhead[C]{}
%   \fancyfoot[L]{%
%     \footerfont%
%     $^{*}$Equal contribution.\\
%     $^{\dagger}$Work done during internship at AIR, Tsinghua University.\\
%     $^{\ddagger}$Corresponding author: Ting Cao (\href{mailto:tingcao@mail.tsinghua.edu.cn}{tingcao@mail.tsinghua.edu.cn}).%
%   }
%   \fancyfoot[R]{}
%   \fancyfoot[C]{}
% }

% The thuair class captures the abstract environment before \maketitle.
% !TEX root = ../neurips_2026.tex
% ===========================================================================
% ABSTRACT
% ===========================================================================
\begin{abstract}
Generative robot policies fail unpredictably at deployment: they hesitate at critical moments, drift off-task, or commit to unrecoverable actions. Existing online failure detectors either require white-box access to policy internals or add runtime overhead through resampling and observation-side signals. Our empirical analysis shows that emitted action chunks themselves already carry strong predictive signal for impending failures in generative robot policies. Motivated by this observation, we introduce \nickname, a lightweight, pure action-space detector that uses two compact signals available from a single forward pass: Temporal Consistency Error (TCE) between consecutive action chunks and Action Chunk Magnitude (ACM) of the current chunk. \nickname maps these signals to per-step failure probabilities with a task-conditioned LSTM--MLP architecture. Across a diverse suite of generative robot policies and benchmarks, \nickname raises alerts before failures become visually recognizable, improving the accuracy (F1)--timeliness Pareto frontier of failure detection by an average hypervolume gain of $+12.7\%$ over both internal- and external-feature baselines, with a $+9.0\%$ early-detection ROC-AUC lead on unseen tasks. \nickname further transfers to deployment, predicting failures on unseen real-robot pick tasks and accelerating RL fine-tuning (PPO) with $2.9\times$ fewer environment interactions.
\end{abstract}

\maketitle

% Teaser figure inlined on the first page (between abstract and introduction)
% in arxiv mode. In NeurIPS mode it floats inside tex/introduction.tex instead.
\begingroup
\captionsetup{font=small,skip=3pt,hypcap=false}
\begin{center}
  \centering
  \includegraphics[trim=0 0 100 160,clip,width=\linewidth]{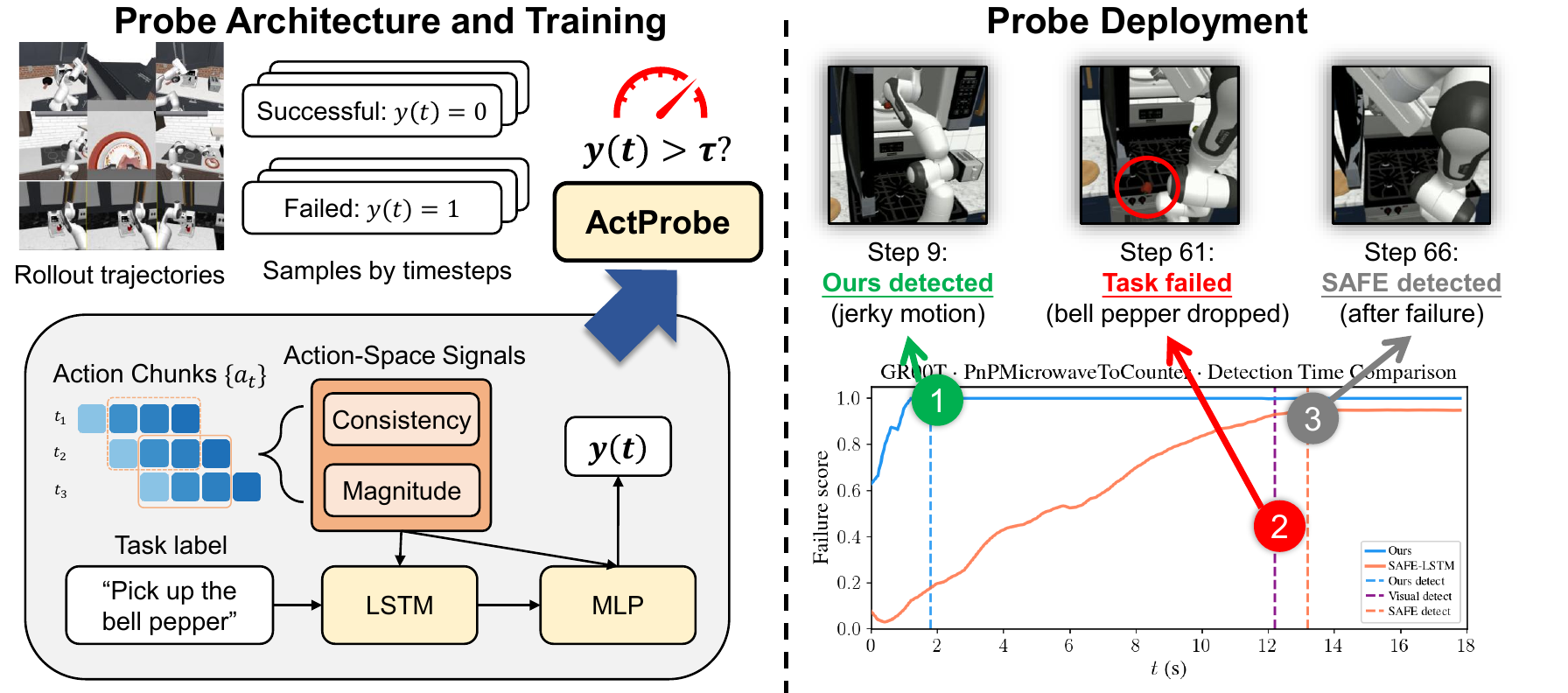}
  \captionof{figure}{Overview of \nickname. \nickname raises alarm before visually recognizable failures by probing action-space signals at runtime. In contrast, SAFE~\cite{gu2025safe}, our strongest baseline, cannot detect failure until the actual failure.}
  \label{fig:teaser}
\end{center}
\endgroup

% !TEX root = ../neurips_2026.tex
% ===========================================================================
% 1. INTRODUCTION
% ===========================================================================
\section{Introduction}
\label{sec:intro}

Generative robot policies, including Vision-Language-Action models (VLAs)~\cite{brohan2022rt1,brohan2023rt2,kim2024openvla,black2024pi0,black2025pi05,bjorck2025groot,li2024cogact} and World-Action Models (WAMs)~\cite{dreamzero2026,nvidia2025cosmos,cen2025rynnvla,bi2025motus}, have emerged as a general-purpose paradigm for end-to-end robot control, demonstrating impressive capabilities across manipulation tasks~\cite{liu2023libero,nasiriany2024robocasa,khazatsky2024droid}.
However, even the strongest generative robot policies fail unpredictably when deployed: they hesitate at critical moments, drift off-task, or commit to unrecoverable actions~\cite{gu2025safe,agia2024stac,farid2021task}.
Reliable deployment therefore demands a complementary capability: failure detection, which lets a system know when its own policy is going wrong, ideally early enough to trigger a fallback or request human intervention~\cite{liu2024fleet,sinha2024realtime,gokmen2023value,du2023vlmsuccess,duan2024aha}.

% --- Fig.1: Teaser ---
% In arxiv mode the teaser is rendered inline right after \maketitle (in main_arxiv.tex)
% so it lands on the first page after the abstract. In NeurIPS mode it floats here.
\ifdefined\arxivversion
% intentionally empty in arxiv mode
\else
\begin{figure}[!t]
  \centering
  \includegraphics[trim = 0 0 100 160,clip,width=\linewidth]{figures/design/teaser.pdf}
  \caption{Overview of \nickname. \nickname is able to raise alarm before visually recognizable failures by probing action-space signals at runtime. In contrast, SAFE~\cite{gu2025safe}, our strongest baseline, cannot detect failure until the actual failure.}
  \label{fig:teaser}
\end{figure}
\fi

A common recipe~\cite{gu2025safe,agia2024stac,xu2025logpzo,romer2025fiper} deploys a lightweight monitor alongside the policy and raises an alert when runtime signals deviate from successful-rollout behavior. Existing detectors fall into two categories based on their input features.
\textbf{(a) Internal-feature} methods~\cite{gu2025safe,francis2026temporal} probe the policy's hidden states. They require white-box access, and these deeply-fused features mix scene appearance with policy behavior, leading the probe to overfit trained tasks (Tab.~\ref{tab:main}) and degrade on unseen ones.
\textbf{(b) External-feature} methods~\cite{agia2024stac,romer2025fiper,lee2025diffdagger,xu2025logpzo} consume the policy's visual observations and emitted action chunks. They typically pair an action-side score with an observation-side or sampling-based component, even though prior work has noted that visual features contribute marginally beyond low-dimensional signals in such detectors~\cite{xu2025logpzo}.

\begin{figure}[t]
  \centering
  \includegraphics[width=0.75\linewidth]{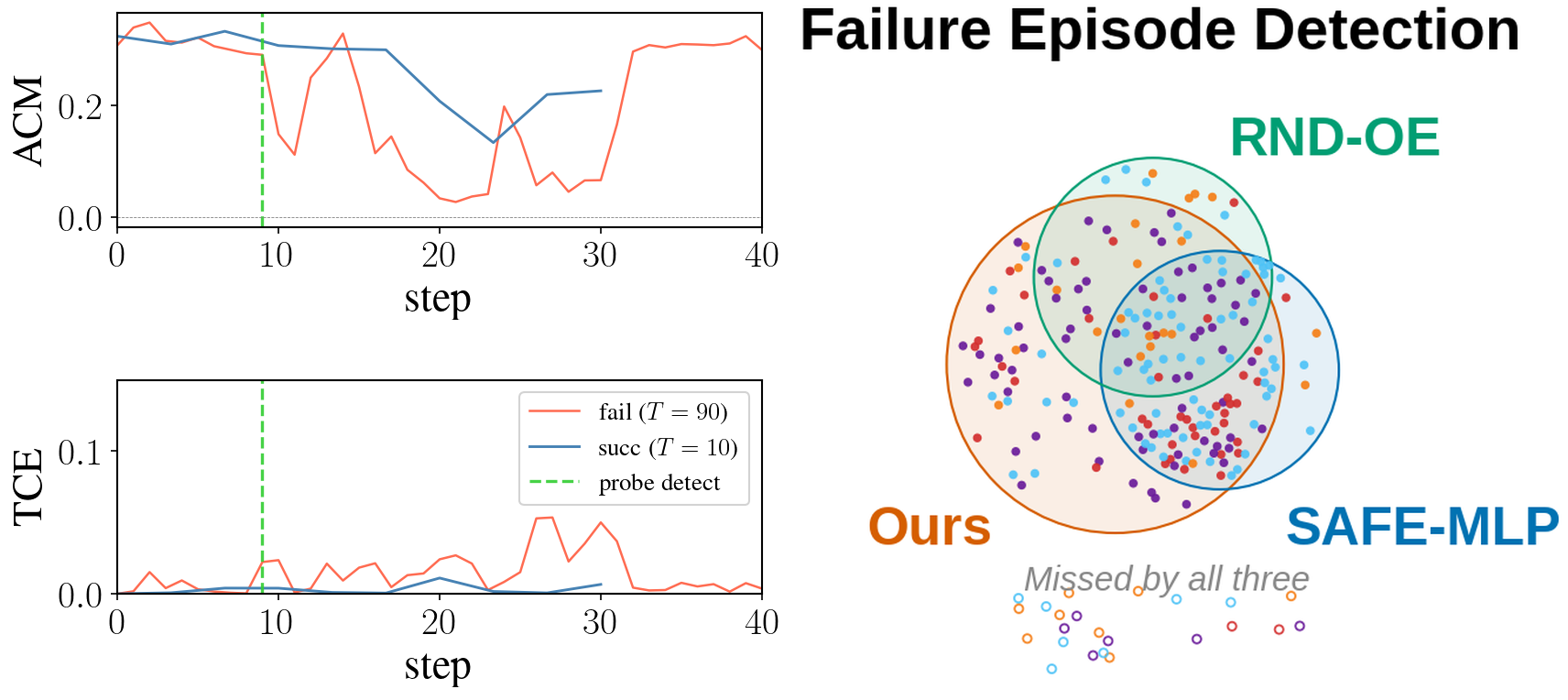}
  \caption{\textbf{Action-space signals distinguish failure from success, and our method subsumes the detection coverage of existing detectors.} \emph{Left}: ACM and TCE trajectories for a representative failure episode (red) and success episode (blue); both signals diverge notably during failure. \emph{Right}: Venn diagram of failure episode detection on $\pi_{0.5}$+RoboCasa; each dot is one failure episode. Our method (Ours) largely encompasses the episodes detected by SAFE-MLP and RND-OE (the RND-based observation-embedding detector from FIPER), while additionally capturing failures they miss.}
  \label{fig:feature_intuition}
\end{figure}

These observations motivate a simpler question: \emph{are the emitted action chunks themselves sufficient for early failure detection in generative robot policies?}
We answer yes, yielding the first \emph{pure action-space} detector: no observation-side fusion, no resampling, and no internal access.
We identify two characteristics of action chunks that are especially informative, and directly available from a single forward pass (Fig.~\ref{fig:feature_intuition}):
\textbf{(a) Temporal Consistency Error (TCE)} between action chunks, measured by the mean squared error (MSE) of their overlapping parts. TCE reflects whether the policy preserves its short-horizon plan across consecutive inferences, serving as a single-pass surrogate for the trajectory jitter that resampling-based methods elicit explicitly.
\textbf{(b) Action Chunk Magnitude (ACM)} of the current action output, measured by its L2 norm. ACM reflects the scale of commands the policy is willing to execute at the current step, distinguishing the small, smooth motions that characterize normal task progress from the oversized corrections that often precede failure, drawing on classical motor-control evidence that purposive motion minimizes magnitude-based smoothness criteria~\cite{flash1985coordination,balasubramanian2015smoothness}.

Building on these features, we propose \nickname, a lightweight yet generalizable early failure detector with two additional design choices.
\nickname adopts a bridged LSTM-MLP architecture that captures both the instantaneous and integrated patterns of action-chunk behavior.
For multi-task training and generalization to unseen tasks, the recurrence is conditioned on a task-label embedding encoded from the original task description.

We evaluate \nickname across five policy--environment settings spanning OpenVLA~\cite{kim2024openvla}, $\pi_0$~\cite{black2024pi0}, $\pi_{0.5}$~\cite{black2025pi05}, and GR00T~\cite{bjorck2025groot} on LIBERO~\cite{liu2023libero} and RoboCasa~\cite{nasiriany2024robocasa}, under seen-task and unseen-task protocols.
Compared to internal- and external-feature baselines, \nickname improves the detection accuracy (F1)--timeliness Pareto frontier with an average hypervolume~\cite{zitzler2003performance} gain of $+12.7\%$ across benchmarks, and outperforms SAFE-MLP, the strongest overall baseline, on unseen tasks by $+9.0\%$ in average early-stage ROC-AUC (\S\ref{sec:early_detection_new}), setting a new state-of-the-art for early failure detection.
We deploy \nickname on both real-robot and downstream simulation use cases (\S\ref{sec:experiment_discard}), demonstrating its practical effectiveness. Specifically, \nickname reaches baseline-level performance with $2.9\times$ fewer environment interactions in RL fine-tuning (PPO) on LIBERO-Object with RLinf~\cite{yu2025rlinf}.
Besides, \nickname is able to transfer across tasks and policies in zero-shot (Table~\ref{tab:cross_category}), and scales effectively with training data (Fig.~\ref{fig:cross_task} and \ref{fig:scaling_more_data}).

We summarize our contributions below:

\begin{itemize}[leftmargin=1.2em,topsep=2pt,itemsep=1pt]
  \item \textbf{Probe feature design.} We show that two action-space signals from a single forward pass—TCE and ACM—suffice for early failure detection, sidestepping both the unseen-task overfitting of internal-feature probes and the runtime overhead of observation- or resampling-based external ones.
  \item \textbf{Probe model design.} We propose \nickname, a lightweight bridged LSTM–MLP architecture with task-embedding conditioning that learns both instantaneous anomalies and integrated behavioral drift from action-space signals.
  \item \textbf{Implementation and evaluation.} We instantiate and evaluate \nickname across five policy--environment settings spanning OpenVLA, $\pi_0$, $\pi_{0.5}$, and GR00T on LIBERO and RoboCasa, yielding a $+12.7\%$ average hypervolume gain on the F1–timeliness frontier and a $+9.0\%$ early-detection ROC-AUC lead ($q{=}0.25$) on unseen tasks.
  \item \textbf{Applications.} We deploy and evaluate \nickname on both real-world and downstream simulation use cases. Specifically, \nickname reaches baseline-level performance with $2.9\times$ fewer environment interactions in RL fine-tuning (PPO) on LIBERO-Object with RLinf.
\end{itemize}
% !TEX root = ../neurips_2026.tex
% ===========================================================================
% 2. RELATED WORK
% ===========================================================================
\section{Related Works}
\label{sec:related}

\subsection{Generative Robot Policies and Action Chunking}

Generative robot policies have emerged as a promising paradigm for robotic manipulation, with two mainstream families.
Vision-Language-Action models (VLAs)~\cite{brohan2022rt1,brohan2023rt2,kim2024openvla,black2024pi0,black2025pi05,bjorck2025groot,li2024cogact,bu2025univla,pertsch2025fast} leverage pretrained vision-language backbones to predict robot actions from visual observations and language instructions.
World-Action Models (WAMs)~\cite{dreamzero2026,nvidia2025cosmos,cen2025rynnvla,bi2025motus,du2023unipi,yang2024unisim} jointly model future observations and actions, conditioning a video world model on the task and rolling out actions consistent with the predicted dynamics.
Unlike early autoregressive policies which predict action tokens one-by-one~\cite{brohan2022rt1,brohan2023rt2,kim2024openvla,pertsch2025fast}, mainstream diffusion or flow-based architectures~\cite{chi2023diffusionpolicy,black2024pi0,bjorck2025groot,li2024cogact} generate a chunk of actions (an ``action chunk''~\cite{zhao2023aloha}) at each inference step, improving temporal consistency of actions.
Concretely, each forward pass emits a horizon of $H$ consecutive actions, and chunks are queried with stride $s\leq H$ (depending on specific configurations), so there are $H-s$ overlapping timesteps between two adjacent action chunks.
\nickname calculates ACM from the last action chunk, and TCE from overlapping steps of the last two action chunks.

\subsection{Failure Detection for Generative Robot Policies}

\paragraph{Internal-feature detectors.}
This line of work builds on the observation that internal policy features can capture high-level information about task success and failure, paralleling evidence that internal representations of language models encode truthfulness and known-unknown signals~\cite{azaria2023internal,kadavath2022know,kuhn2023semantic}.
As demonstrated by \citet{gu2025safe}, straightforward statistics in feature space (e.g., Mahalanobis distance, $k$-NN) can already perform well on a few benchmarks, echoing classical feature-space OOD detectors in supervised learning~\cite{hendrycks2017baseline,lakshminarayanan2017ensembles,gal2016dropout}.
With a learning-based probe (e.g., LSTM or MLP), SAFE~\cite{gu2025safe} and SAFE-TDQC~\cite{francis2026temporal} further improve prediction accuracy, setting a new state-of-the-art among internal-feature detectors.
However, deeply-fused policy features entangle scene appearance with policy behavior, which we find are hard to learn by a lightweight probe and limit cross-task transfer (\S\ref{sec:experiments}); they also require white-box access to the policy, which is unavailable for closed-source models.

\paragraph{External-feature detectors.}
A second line of work consumes only what the policy exposes externally: visual observations, task commands, and emitted action chunks. Existing detectors pair an action-side score with an observation- or sampling-based component, including STAC~\cite{agia2024stac}, FIPER~\cite{romer2025fiper}, and per-sample denoising-variance estimators~\cite{lee2025diffdagger,he2025uncertaintyfree}; LogpZO~\cite{xu2025logpzo} sits adjacent to this thread, fitting an out-of-distribution density to a learned embedding rather than to raw actions.
A complementary thread uses vision-language models as external success or failure verifiers~\cite{du2023vlmsuccess,duan2024aha}, and statistical task-driven OOD detectors provide distribution-shift guarantees for robot learning more broadly~\cite{farid2021task,liu2024fleet,sinha2024realtime}.
Among these signals, prior work has noted that visual features tend to contribute less than low-dimensional ones~\cite{xu2025logpzo}.
\nickname therefore operates purely in the action space, to the best of our knowledge the first such detector. It uses no observation-side fusion and no resampling, relying on two scalars (TCE, ACM) extracted from the existing single forward pass and trained jointly with a task-label embedding.

% !TEX root = ../neurips_2026.tex
% ===========================================================================
% 3. METHOD
% ===========================================================================
\section{Method}
\label{sec:method}

\subsection{Action-Space Signals from Generative Robot Policies}
\label{sec:features}

\paragraph{Action chunks.}
We consider a generative robot policy $\pi$ that, at decision step $t$, observes $o_t$ (comprising the language prompt, visual observations, and proprioceptive robot state) and samples an action chunk~\cite{zhao2023aloha,chi2023diffusionpolicy}
\[
  a_t \;\sim\; \pi(\cdot \mid o_t), \qquad a_t \;:=\; \bigl(a_{t \mid t},\, a_{t+1 \mid t},\, \ldots,\, a_{t+H-1 \mid t}\bigr),
\]
where $a_{t+i \mid t} \in \mathbb{R}^{D_a}$ is the action prediction for physical timestep $t+i$ made at decision step $t$, $D_a$ is the action dimension, and $H$ is the chunk horizon. The first $s \le H$ actions are executed before the policy re-evaluates and emits $a_{t+s}$, leaving $H-s$ overlapping timesteps between adjacent chunks. Autoregressive policies such as OpenVLA~\cite{kim2024openvla} correspond to the degenerate setting $H=s=1$.

From $a_t$ we extract two scalar features per decision step, Temporal Consistency Error (TCE) and Action Chunk Magnitude (ACM), both directly available without resampling, auxiliary forward passes, or access to internal policy states.

\paragraph{Temporal Consistency Error (TCE).}
TCE measures whether the policy preserves its short-horizon plan across consecutive inferences, by comparing the overlapping parts of two consecutive action chunks $a_{t-s}$ and $a_t$ via the mean squared error,
\[
  \mathrm{TCE}_t \;=\; \frac{1}{(H-s)\,D_a}\sum_{i=0}^{H-s-1}\bigl\| a_{t+i \mid t-s} - a_{t+i \mid t} \bigr\|_2^2 .
\]
Persistently large $\mathrm{TCE}_t$ indicates that the policy is repeatedly revising its near-future plan rather than committing to a stable trajectory, a single-pass surrogate for the trajectory jitter that resampling-based methods elicit explicitly~\cite{agia2024stac,romer2025fiper,lee2025diffdagger}. For autoregressive policies, where adjacent inferences have no native overlap, TCE degenerates to the squared difference between consecutive single-step actions, $\mathrm{TCE}_t = \|a_t - a_{t-1}\|_2^2 / D_a$.

\paragraph{Action Chunk Magnitude (ACM).}
ACM is the L2 norm of the current action chunk treated as a flat vector,
\[
  \mathrm{ACM}_t \;=\; \bigl\| a_t \bigr\|_2 \;=\; \sqrt{\sum_{i=0}^{H-1} \bigl\| a_{t+i \mid t} \bigr\|_2^2}.
\]
ACM gives the probe a coarse description of the magnitude of commands the policy outputs at the current step. It distinguishes the small, smooth motions that characterize normal task progress, a property long observed in human and learned manipulation~\cite{flash1985coordination,balasubramanian2015smoothness,zhao2023aloha,chi2023diffusionpolicy}, from the oversized corrections that often precede failure.

\subsection{Model Architecture and Training}
\label{sec:architecture}

\paragraph{Bridged LSTM--MLP architecture.}
\nickname maps the feature sequence to a failure probability at each step, $p_t \in [0,1]$. Let $f_t = (\mathrm{TCE}_t,\, \mathrm{ACM}_t) \in \mathbb{R}^2$ denote the feature vector at step $t$. We standardize (z-score) the features and append the normalized timestep $\tau_t = t/(T{-}1) \in [0,1]$, where $T$ denotes the episode timeout in action-chunk steps, to form the input $x_t = [\tilde f_t;\, \tau_t]$. A single-layer LSTM processes the sequence,
\[
  (h_t, c_t) \;=\; \mathrm{LSTM}(x_t,\, h_{t-1},\, c_{t-1}),
\]
and a small MLP, fed with the concatenation of the LSTM hidden state and the raw input, produces the corresponding probability,
\[
  p_t \;=\; \sigma\!\bigl(\mathrm{MLP}([h_t;\, x_t])\bigr).
\]
The two pathways play complementary roles: the LSTM aggregates long-range behavioral history, capturing the \emph{integrated} drift of TCE and ACM across an episode, while the \emph{bridge}, a skip connection that carries the raw input $x_t$ straight to the MLP, gives it direct access to the \emph{instantaneous} feature values at step $t$. This lets \nickname respond to both gradual behavioral drift and abrupt single-step anomalies with a probe of only $\sim$24K parameters. Full architectural specifications are in Appendix~\ref{app:training}; ablation studies are in \S\ref{sec:two_feature_signal_new} and Appendix~\ref{app:ablations}.

\paragraph{Task-label conditioning.}
To enable multi-task training and generalization to unseen tasks, the LSTM is conditioned on a \emph{task-label embedding} obtained from the original task description. For each task, the instruction $\ell$ is encoded offline once by a frozen 0.6B-parameter Qwen3-Embedding model~\cite{qwen3embedding2025} and projected through a small bottleneck to initialize the recurrent state, $(h_0, c_0) = \mathrm{LangProj}(\ell)$. Conditioning is injected only at the recurrence's initial state rather than concatenated to every $x_t$, which keeps the per-step feature dimension at 3 and pushes the probe toward learning task-conditional thresholds on TCE and ACM instead of memorizing task-specific patterns. In our implementation, the online \nickname inference pipeline adds approximately 3 ms per probe call. As shown in Section~\ref{sec:experiments}, this design transfers more reliably to held-out tasks than the hidden-state probes of prior work.

\paragraph{Training and deployment.}
\nickname operates at the action-chunk rate in both training and deployment: one probe input is formed when the policy emits a new action chunk. During training, the rollout outcome is broadcast to all action-chunk steps in the episode, and the probe is optimized with binary cross-entropy averaged over valid chunk steps. At deployment, the probe outputs a failure probability for each new action chunk; we keep the largest probability seen so far and raise an alert once it exceeds a scalar threshold $\tau$. We set $\tau$ by applying split conformal calibration~\cite{vovk2005algorithmic} to the maximum probe scores of held-out successful rollouts at a target false-positive rate $\alpha$ (the conformal significance level; default $0.15$), so that successful rollouts exceed $\tau$ with probability at most $\alpha$; larger $\alpha$ yields a more aggressive operating point. Compared with the time-varying functional conformal bands used by prior work~\cite{gu2025safe}, this single-threshold trigger is simpler to calibrate and deploy across tasks.

% !TEX root = ../neurips_2026.tex
% ===========================================================================
% 4. EXPERIMENTS
% ===========================================================================
\section{Experiments}
\label{sec:experiments}

We organize the experiments around three questions: how effectively \nickname improves early-detection accuracy and timeliness (\S\ref{sec:early_detection_new}, \S\ref{sec:two_feature_signal_new}, \S\ref{sec:scaling_more_data}); how well its action-space signals generalize across tasks and policies (\S\ref{sec:cross_task_transfer}); and how it supports practical downstream applications (\S\ref{sec:experiment_discard}).

\subsection{Experimental setup}
\label{sec:setup_new}

\paragraph{Benchmarks.}
We evaluate \nickname across five policy--environment settings: two on LIBERO-10 and three on RoboCasa.
On \textbf{LIBERO-10}~\cite{liu2023libero}, a standard long-horizon tabletop manipulation benchmark, we evaluate $\pi_0$~\cite{black2024pi0} and OpenVLA~\cite{kim2024openvla} across 10 long tasks. On \textbf{RoboCasa}~\cite{nasiriany2024robocasa}, a diverse household manipulation benchmark, we evaluate GR00T~\cite{bjorck2025groot} on \emph{single-stage (atomic)} tasks and $\pi_{0.5}$~\cite{black2025pi05} on both \emph{single-stage} and \emph{multi-stage (composite)} task groups. The composite study uses a five-task subset spanning pantry restocking, arrange vegetables, cookware soaking, microwave thawing and coffee preparation. Together, these settings cover the major benchmark families used in recent evaluations of embodied robot policies and span representative autoregressive and diffusion/flow-based policy families.

\paragraph{Baselines.}
We compare against both white-box and black-box failure detectors from prior work. \textbf{(a) White-box baselines} access internal policy representations: Cosine $k$-NN serves as a non-parametric hidden-state baseline, while the SAFE family~\cite{gu2025safe}, which represents the current state of the art in white-box VLA failure detection, learns supervised detectors over hidden-state sequences or per-step embeddings via SAFE-LSTM, SAFE-MLP, and their Temporal-Difference Q-based
Calibration (TDQC) variants~\cite{francis2026temporal}. The TDQC variants retain the same hidden-state inputs and probe backbones but replace standard rollout-success supervision with temporal-difference calibration targets for sequential success prediction. \textbf{(b) Black-box baselines} use only externally exposed policy signals: LogpZO~\cite{xu2025logpzo} follows its original score-based failure-detection formulation, fitting a per-task model over action sequences and triggering when the current log-probability falls outside the calibration distribution. STAC-Single~\cite{agia2024stac} is SAFE's strongest reproduction of STAC, outperforming its resample variant: it uses the policy's native single-sample action chunks and measures disagreement on the overlapping horizon between consecutive chunks, without repeated resampling or a VLM progress module. All trainable baselines are trained on the same rollout sets as \nickname.

\paragraph{Metrics.}
\textbf{(a) F1 and $T\text{-det}$ given different alarm thresholds.}
For deployment-oriented evaluation, we measure detection accuracy (F1) and timeliness (the normalized first-alert time $T\text{-det}$) under different alarm thresholds, each set by the target false-positive rate $\alpha$ used for calibration in \S\ref{sec:architecture}. Sweeping the threshold $\alpha$ from $0.02$ to $0.95$ traces an accuracy--timeliness curve per detector, which we summarize by its hypervolume~\cite{zitzler2003performance}---the area dominated by the curve relative to the reference point $(T\text{-det}{=}1,\ \text{F1}{=}0)$. For \nickname, $\alpha$ sets a single constant cutoff $\tau(\alpha)$ calibrated on validation successes; for prior detectors, $\alpha$ indexes each method's native functional conformal prediction (FunctionalCP) band~\cite{gu2025safe,francis2026temporal}, a time-varying cutoff calibrated to keep the false-positive rate at most $\alpha$.
\textbf{(b) ROC-AUC of detectors at certain timesteps.}
Independent of any calibration or threshold, this metric summarizes how well a detector separates failed from successful rollouts, isolating signal quality. We compute it from the maximum score attained within the first 
$q$ fraction of each rollout, using the first quarter ($q{=}0.25$) as our primary early-detection setting. 

\begin{figure}[!t]
  \centering
  \includegraphics[width=0.8\linewidth]{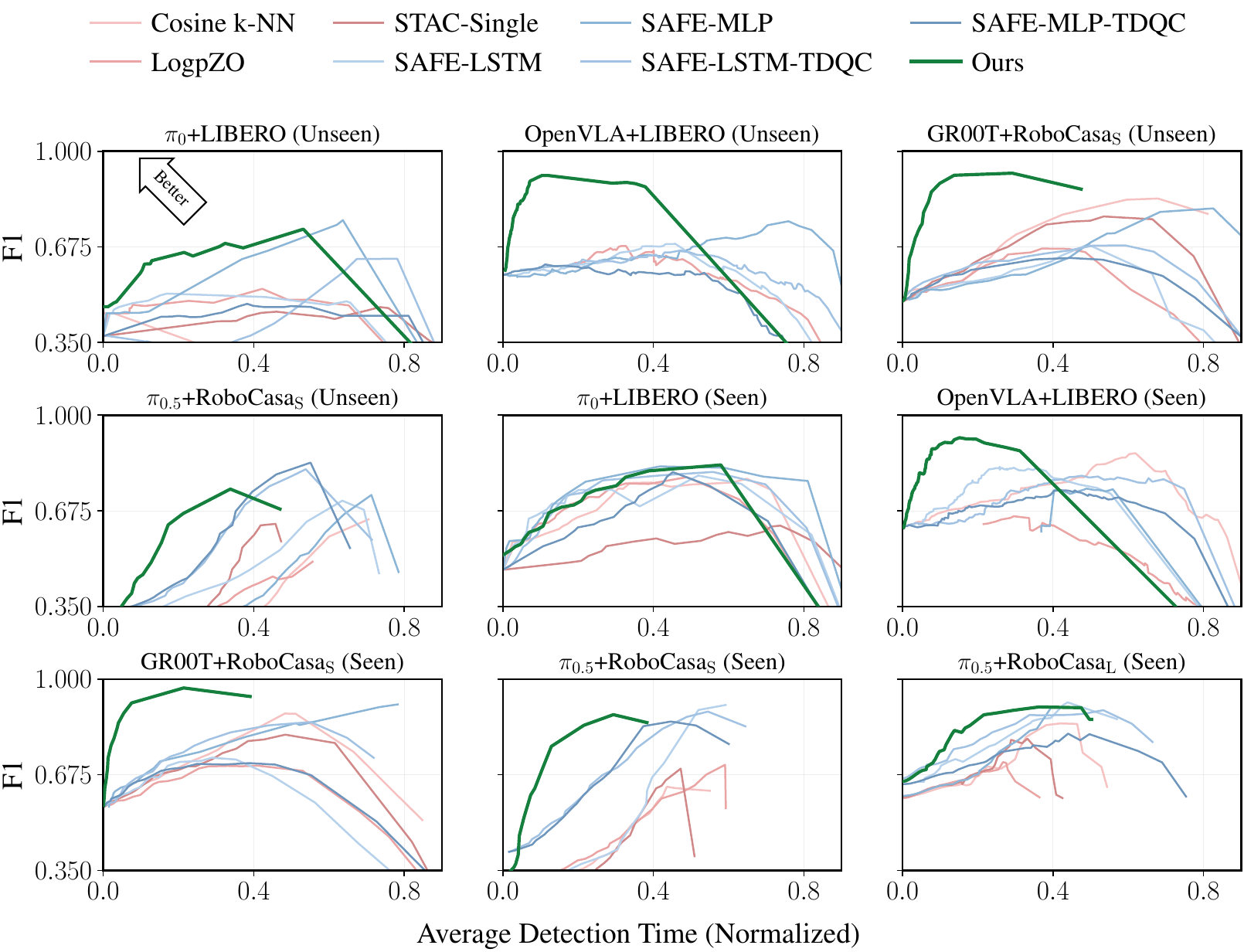}
  \caption{\textbf{F1 vs.\ average detection time ($T$-det) across benchmarks and protocols.} Each panel title indicates the benchmark and protocol (Seen/Unseen). RoboCasa$_{\mathrm{S}}$ and RoboCasa$_{\mathrm{L}}$ denote single-stage and multi-stage task groups, respectively. Points toward the top-left indicate more accurate \emph{and} earlier detection. \nickname (ours) places curves toward the top-left in most benchmarks under both protocols.}
  \label{fig:early_detection}
\end{figure}

\begin{table}[!t]
  \caption{\textbf{Early-detection AUC across five benchmarks} ($q{=}0.25$: ROC-AUC over the first quarter of the rollout), averaged over 3 seeds. RoboCasa$_{\mathrm{S}}$ and RoboCasa$_{\mathrm{L}}$ denote the single-stage (atomic) and multi-stage (composite) task groups, respectively. ``---'' indicates the method does not apply (STAC-Single is omitted on OpenVLA, which commits to a single action per inference). \textbf{Bold}: best per column; \underline{underlined}: second best. The shaded row marks our method (\nickname).}
  \label{tab:main}
  \centering
  \small
  \setlength{\tabcolsep}{2.5pt}
  \renewcommand{\arraystretch}{0.95}
  \begin{tabular}{l|cc|cc|cc|cc|c}
    \toprule
    & \multicolumn{2}{c|}{\shortstack{$\pi_0$ \\ +LIBERO}} & \multicolumn{2}{c|}{\shortstack{OpenVLA \\ +LIBERO}} & \multicolumn{2}{c|}{\shortstack{GR00T \\ +RoboCasa}} & \multicolumn{2}{c|}{\shortstack{$\pi_{0.5}$ \\ +RoboCasa$_{\mathrm{S}}$}} & \shortstack{$\pi_{0.5}$ \\ +RoboCasa$_{\mathrm{L}}$} \\
    Method & Seen & Unseen & Seen & Unseen & Seen & Unseen & Seen & Unseen & Seen \\
    \midrule
    Cosine $k$-NN      & 68.9 & 57.2 & 72.1 & 42.0 & 65.5 & 70.3 & 56.3 & 44.4 & 71.1 \\
    \midrule
    LogpZO             & 73.1 & 58.3 & 57.7 & 52.6 & 68.6 & 65.5 & 70.7 & 67.2 & 73.8 \\
    STAC-Single        & 65.3 & \underline{65.2} & --- & --- & 73.6 & \underline{73.8} & 62.3 & 66.0 & 77.7 \\
    \midrule
    SAFE-LSTM          & 78.2 & 59.2 & 75.7 & 55.9 & 70.9 & 61.3 & \textbf{74.9} & \underline{73.1} & \underline{80.9} \\
    SAFE-MLP           & 77.3 & 58.9 & \textbf{80.9} & \underline{67.9} & 74.2 & 61.5 & 73.7 & 71.0 & 74.7 \\
    SAFE-LSTM-TDQC     & \underline{79.6} & 62.8 & 67.5 & 57.6 & \underline{74.2} & 65.9 & 73.9 & 62.4 & 76.1 \\
    SAFE-MLP-TDQC      & 76.6 & 60.7 & 65.7 & 52.8 & 71.3 & 64.7 & \underline{74.8} & 70.4 & 73.7 \\
    \midrule
    \rowcolor{gray!15}
    \textbf{\nickname} & \textbf{80.1} & \textbf{71.4} & \underline{76.3} & \textbf{73.8} & \textbf{77.4} & \textbf{76.3} & 70.0 & \textbf{73.8} & \textbf{82.7} \\
    \bottomrule
  \end{tabular}
\end{table}

\subsection{End-to-End Comparison}
\label{sec:early_detection_new}

We evaluate the accuracy–timeliness trade-off from two complementary angles: the F1 vs.\ $T$-det curves (Figure~\ref{fig:early_detection}), where curves toward the top-left indicate more accurate and earlier detection, and the early-detection AUC at $q{=}0.25$ (Table~\ref{tab:main}). All methods are trained on the same 50 episodes per task.

\textbf{(a) \nickname advances the Pareto frontier of detection accuracy (F1) vs.\ detection time (Figure~\ref{fig:early_detection}) on most benchmarks}, with an average hypervolume gain of $+12.7\%$ across benchmarks. These curves are also directly actionable at deployment: given an admissible false-alarm rate (set by $\alpha$) and a maximum tolerable detection latency, the recommended detector is the one attaining the highest F1 within that region.

% \ifdefined\arxivversion\FloatBarrier\fi

\textbf{(b) \nickname also improves early-stage detection ($q{=}0.25$ ROC-AUC, Table~\ref{tab:main}).} \nickname averages $75.8\%$ across nine splits, $+4.6\%$ above the strongest baseline (SAFE-MLP) and widening to $+9.0\%$ on unseen tasks. On seen tasks, hidden-state probes stay competitive in this limited-data regime, yet \nickname matches them there with no white-box access. Overall, \nickname{} generalizes better than both internal-feature probes (which overfit seen tasks) and external-feature detectors that resample or rely on observations, with the advantage most pronounced on unseen tasks.

\subsection{Ablation Study}
\label{sec:two_feature_signal_new}

This ablation studies two design choices behind \nickname.
\textbf{The computation of the temporal consistency signal.} TCE measures the discrepancy between the overlapping parts of consecutive action chunks, which we instantiate as MSE. Prior detectors instead elicit chunk disagreement by resampling $K$ alternative chunks; we therefore vary both the signal's source (native overlap vs.\ $K$-resampling) and the distance (MSE, MMD \cite{agia2024stac}, ACE \cite{romer2025fiper}) to identify the most discriminative combination.
\textbf{The contribution of the bridged LSTM--MLP architecture.} \nickname maps these signals to a failure score with an LSTM--MLP designed to capture both long-range behavioral trends and instantaneous changes; we ablate each component to test whether coupling the two is responsible for the gain.

\begin{table*}[!t]
\centering
\caption{\textbf{Distance $\times$ architecture ablation on $\pi_{0.5}$+RoboCasa ($q{=}0.25$ ROC-AUC, \%).} Averaged over 3 seeds. Seen uses 24 training tasks; unseen uses 7 held-out tasks. Rows progress through increasingly structured variants: raw K-resampling distance without learning, K-resample+MLP, overlap+MLP, overlap+LSTM, and finally \nickname.}
\label{tab:ablation_distance_arch}
\small
\setlength{\tabcolsep}{5pt}
\begin{tabular}{lccc|ccc}
\toprule
\multicolumn{1}{c}{Variant} & \multicolumn{3}{c|}{Seen} & \multicolumn{3}{c}{Unseen} \\
\cmidrule(lr){2-4}\cmidrule(lr){5-7}
 & ACE & MMD & MSE & ACE & MMD & MSE \\
\midrule
K-resample (no learn)
& 58.7 & 56.7 & 61.1
& 60.8 & 52.3 & 68.4 \\
K-resample + MLP
& \textbf{63.0} & 61.7 & 62.5
& 60.6 & 57.5 & 68.2 \\
overlap + MLP
& 60.1 & 64.9 & 64.3
& 60.7 & 62.3 & 70.5 \\
overlap + LSTM
& 59.2 & 63.9 & 68.1
& 59.8 & 63.3 & 69.6 \\
\nickname
& 60.7 & \textbf{67.6} & \textbf{69.7}
& \textbf{62.7} & \textbf{66.3} & \textbf{72.1} \\
FIPER & 67.0 & --- & --- & 71.0 & --- & --- \\
\bottomrule
\end{tabular}
\end{table*}

Three results stand out in Table~\ref{tab:ablation_distance_arch}. \textbf{(a) MSE preserves more action-space state information than ACE or MMD:} across all five variants, it yields the highest AUC on unseen tasks and ties or wins on seen tasks. \textbf{(b) Native overlap between consecutive replans yields a more discriminative signal than resampling-based uncertainty:} native overlap outperforms both the statistical and learned K-resampling variants, and our resampling-based FIPER reproduction stays below \nickname{}. \textbf{(c) The bridged LSTM--MLP improves early-detection AUC over an MLP-only mapping:} adding an LSTM over the overlap signal already improves on the MLP-only mapping, and the full bridged LSTM--MLP improves it further---a total $+5.4\%$ over overlap+MLP on MSE-seen---where the recurrent state accumulates evidence across replans while the MLP path preserves sensitivity to the current chunk.

\subsection{Cross-Task Transfer}
\label{sec:cross_task_transfer}

\begin{figure}[!t]
  \centering
  \includegraphics[width=\linewidth]{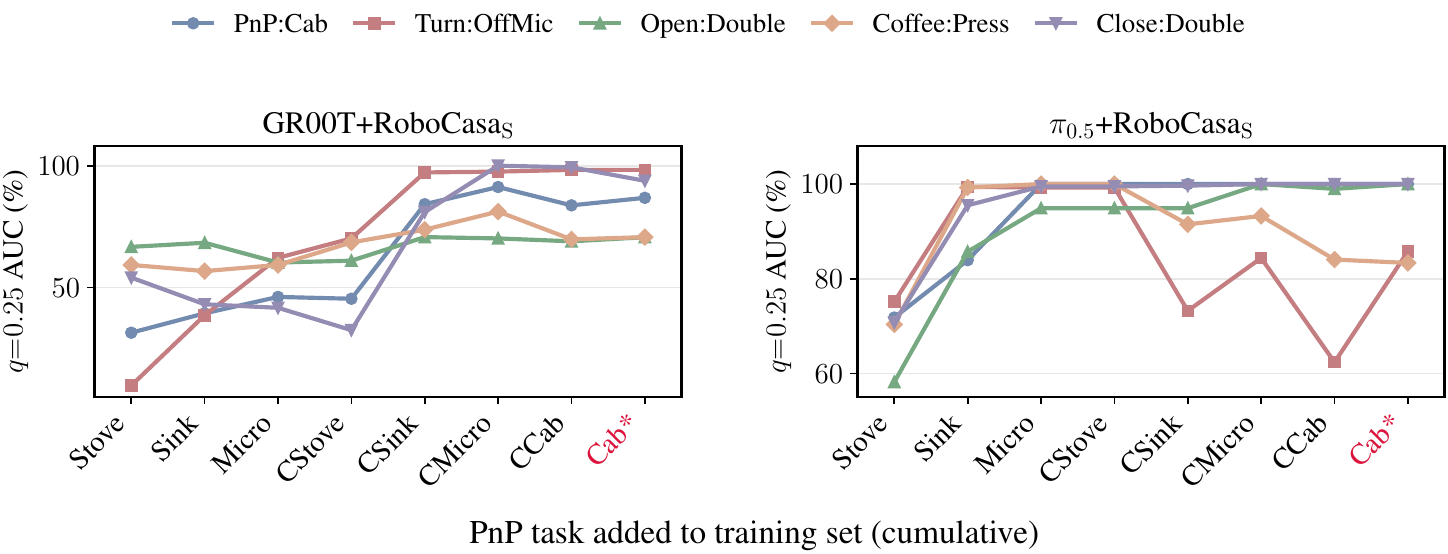}
  \caption{\textbf{\nickname trained on PnP generalizes across categories as training data grows.} Left: GR00T+RoboCasa. Right: $\pi_{0.5}$+RoboCasa. Each curve tracks one of five held-out test tasks as PnP training tasks are added cumulatively (x-axis). Tasks marked with~$*$ (red) denote the corresponding held-out target tasks. In both settings, \nickname reaches high detection AUC once a few behaviorally rich tasks are seen.}
  \label{fig:cross_task}
\end{figure}

We characterize the generalization of \nickname from two angles. \textbf{Task-level transfer} covers cross-category training (Table~\ref{tab:cross_category}, Left), within-family few-shot transfer (Figure~\ref{fig:cross_task}), and language-embedding substitution at test time (Table~\ref{tab:lang_pert}). \textbf{Policy-level transfer} covers zero-shot deployment of the GR00T-trained \nickname{} on $\pi_{0.5}$+RoboCasa (Table~\ref{tab:cross_category}, Right).

\begin{table}[!t]
\centering
\caption{\textbf{Cross-category and cross-model transfer ($q{=}0.25$ AUC, \%).} \emph{Left}: cross-category transfer matrix for GR00T+RoboCasa; each cell reports \nickname trained on the row category and evaluated on unseen tasks of the column category; diagonal entries (gray) use the seen-task protocol. \textbf{Bold}: column-best off-diagonal. \emph{Right}: zero-shot cross-model transfer---\nickname trained on GR00T+RoboCasa and evaluated directly on $\pi_{0.5}$+RoboCasa per category.}
\label{tab:cross_category}
\small
\setlength{\tabcolsep}{3.5pt}
\begin{minipage}[t]{0.47\linewidth}
\centering
\textbf{GR00T+RoboCasa} (3 seeds)\\[3pt]
\begin{tabular}{lccccc}
\toprule
{\scriptsize Train$\downarrow$/Test$\rightarrow$} & PnP & Turn & Open & Coffee & Close \\
\midrule
PnP    & \cellcolor{gray!15}\textit{\textbf{72.5}} & 54.5 & \textbf{70.0} & 62.6 & 53.7 \\
Turn   & 53.3 & \cellcolor{gray!15}\textit{\textbf{81.3}} & 61.8 & \textbf{71.2} & \textbf{62.4} \\
Open   & \textbf{65.8} & \textbf{72.1} & \cellcolor{gray!15}\textit{\textbf{82.9}} & 58.1 & 58.9 \\
Coffee & 62.4 & 66.0 & 46.7 & \cellcolor{gray!15}\textit{\textbf{82.3}} & 58.5 \\
Close  & 61.4 & 70.3 & 38.9 & 58.9 & \cellcolor{gray!15}\textit{\textbf{64.9}} \\
\midrule
SAFE-MLP & 68.5 & 72.5 & 70.4 & 77.4 & 39.6 \\
\bottomrule
\end{tabular}
\end{minipage}\hspace{1em}%
\begin{minipage}[t]{0.40\linewidth}
\centering
GR00T $\to$ $\pi_{0.5}$+RoboCasa (zero-shot)\\[3pt]
\begin{tabular}{lc}
\toprule
Category & $q{=}0.25$ AUC (\%) \\
\midrule
PnP    & 67.1 \\
Turn   & 60.9 \\
Open   & 59.5 \\
Coffee & 58.1 \\
Close  & 69.8 \\
Multi  & 70.2 \\
\bottomrule
\end{tabular}
\end{minipage}
\end{table}

On task-level transfer, \textbf{(a) behaviorally rich categories transfer best across tasks}: in Table~\ref{tab:cross_category}~(Left), PnP (pick-and-place), Turn (turning appliances on/off), and Open (opening doors/drawers) jointly produce all five column-best off-diagonal transfers, we attribute this to their diverse and fine-grained manipulation, which exposes the probe to a broader range of failure patterns. \textbf{(b) Within such a behaviorally rich category, only a few source tasks suffice for cross-category generalization}: Figure~\ref{fig:cross_task} trains \nickname{} on PnP tasks added cumulatively ($50$ episodes per task) and tests on held-out targets spanning all five RoboCasa categories; the full PnP curriculum reaches an average AUC of $0.86$ across GR00T and $\pi_{0.5}$. \textbf{(c) \nickname{}'s generalization draws on both action-space failure patterns and language semantics.} Replacing the correct task embedding with a same-category one (semantically similar but not identical) still lets \nickname beat SAFE-MLP on 4 of 5 categories (Table~\ref{tab:lang_pert}), with up to $+33.7\%$ on Close, showing that semantically related language suffices for strong performance. A semantically unrelated embedding, by contrast, is generally worse than providing no language at all.

\begin{table}[!t]
\centering
\caption{\textbf{Language embedding substitution at test time on GR00T+RoboCasa.} Each row uses an \nickname model trained on that category. \emph{Correct}: task's own embedding. \emph{Same-cat.\ lang}: replaced with another embedding from the same test category. \emph{Cross-cat.\ lang}: replaced with an embedding from a different category. Bold: best per row.}
\label{tab:lang_pert}
\small
\begin{tabular}{lcccccc}
\toprule
Category & SAFE-MLP & \makecell{\nickname\\(no lang)} & Correct & \makecell{Same-cat.\\lang} & \makecell{Cross-cat.\\lang} \\
\midrule
PnP    & 68.5 & 61.2 & \textbf{72.5} & 68.9 & 59.2 \\
Turn   & 72.5 & 77.5 & 81.3 & \textbf{81.6} & 70.9 \\
Open   & 70.4 & 70.6 & \textbf{83.0} & 78.2 & 49.1 \\
Coffee & 77.4 & 62.8 & \textbf{82.3} & 75.2 & 40.0 \\
Close  & 39.6 & 64.4 & 64.9 & \textbf{73.3} & 64.9 \\
\bottomrule
\end{tabular}
\end{table}

On policy-level transfer, \textbf{action-space signals generalize across policies zero-shot}: applying the GR00T-trained \nickname{} directly to $\pi_{0.5}$+RoboCasa (Table~\ref{tab:cross_category}, Right) yields $58.1\%$--$70.2\%$ AUC across six categories, well above chance, confirming that the learned signals capture a policy-agnostic failure signature.

\subsection{Scaling with More Data}
\label{sec:scaling_more_data}

\begin{figure}[!t]
  \centering
  \includegraphics[width=0.95\linewidth]{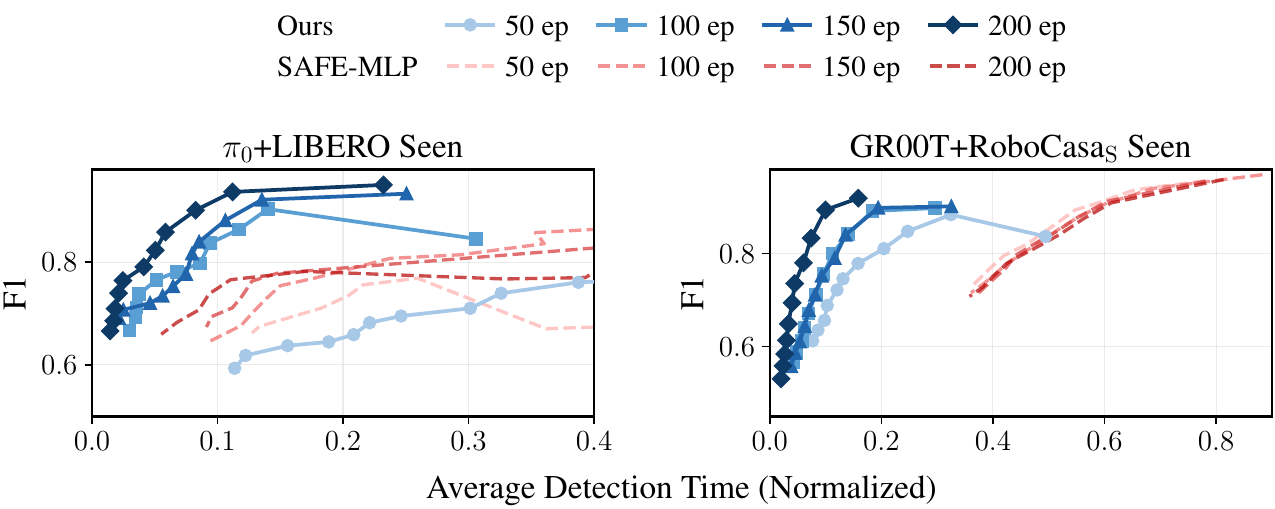}
  \caption{\textbf{\nickname improves steadily as more training rollouts are added.} F1 vs.\ normalized average detection time for the same two-feature probe trained with 50, 100, 150, or 200 episodes per task. Left: $\pi_0$+LIBERO seen-task protocol. Right: GR00T+RoboCasa seen-task protocol. Dashed lines show SAFE-MLP at the same data levels. In both settings, adding more data shifts the \nickname frontier toward the upper left, widening the gap over SAFE-MLP.}
  \label{fig:scaling_more_data}
\end{figure}

We scale the training episodes per task from 50 to 200 on $\pi_0$+LIBERO and GR00T+RoboCasa and compare \nickname{} to SAFE-MLP at each data level (Figure~\ref{fig:scaling_more_data}). \textbf{\nickname{}'s scaling gains substantially outpace SAFE-MLP's.} We attribute this to a difference in what each detector learns from extra data: SAFE-MLP, which separates the hidden-state distributions of success and failure, mainly refines an existing decision boundary, whereas \nickname{} extracts increasingly finer action-space failure signals from each additional rollout, progressively pushing the F1--timeliness Pareto frontier outward. This makes \nickname{} the more promising choice when higher-precision detection is required.

\begin{table}[!t]
  \centering
  \caption{\textbf{Real-robot deployment.} Number of detector-correct trials out of 12 recorded rollouts on two unseen pick tasks. \textbf{Bold}: best per row.}
  \label{tab:rl_robot}
  \small
  \begin{tabular}{lccc}
    \toprule
    Task & \nickname & STAC-Single & SAFE-MLP \\
    \midrule
    Pick Yellow Pear  ($n{=}12$) & \textbf{6/12} & 0/12 & 2/12 \\
    Pick Milk Carton  ($n{=}12$) & \textbf{5/12} & 1/12 & 3/12 \\
    \bottomrule
  \end{tabular}
\end{table}

\begin{figure}[!t]
  \centering
  \begin{subfigure}[b]{0.385\linewidth}
    \centering
    \includegraphics[height=2.3in,keepaspectratio]{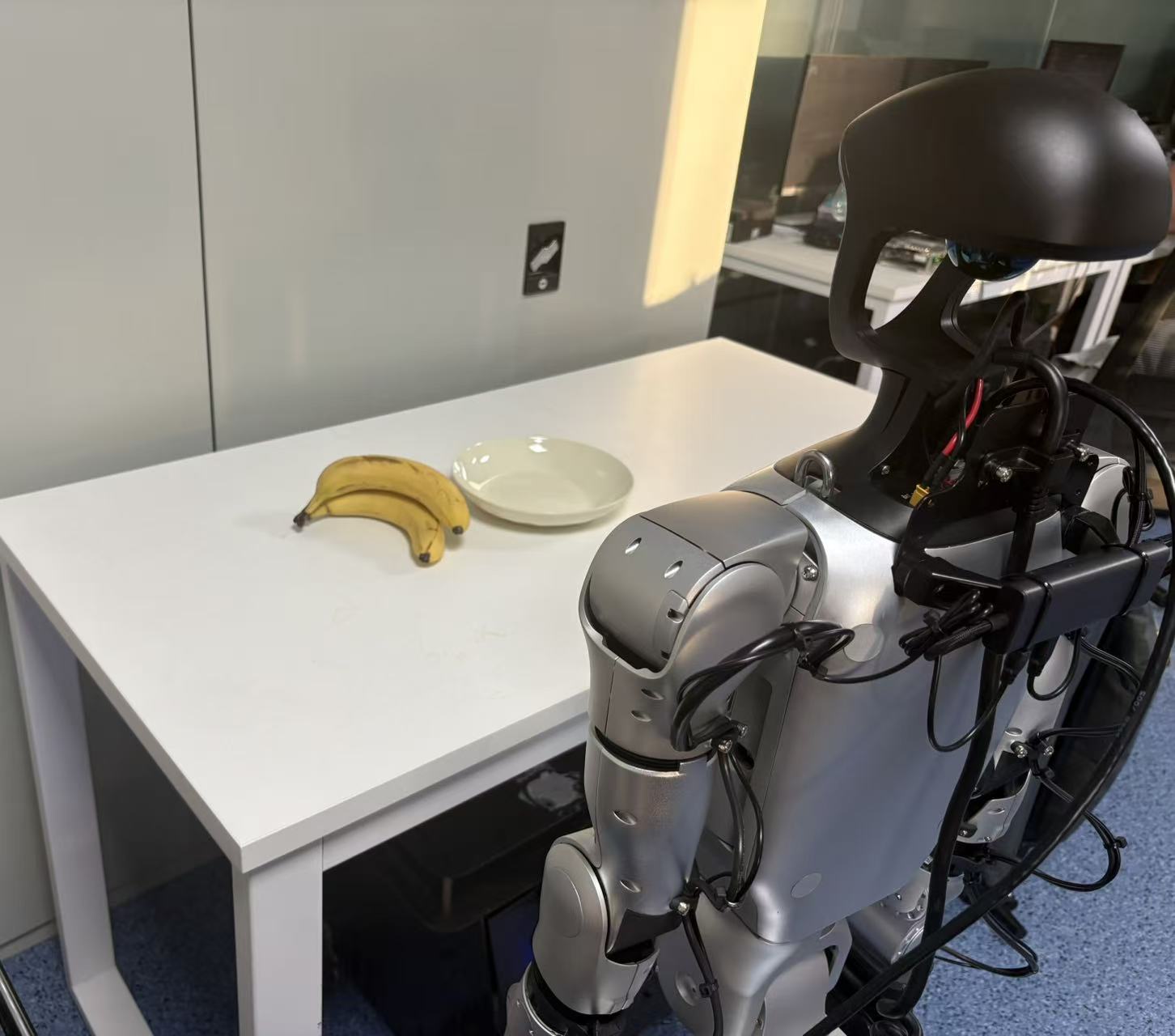}
    \caption{Real-robot setup.}
    \label{fig:robot_setup}
  \end{subfigure}\hspace{1em}%
  \begin{subfigure}[b]{0.466\linewidth}
    \centering
    \includegraphics[height=2.3in,keepaspectratio]{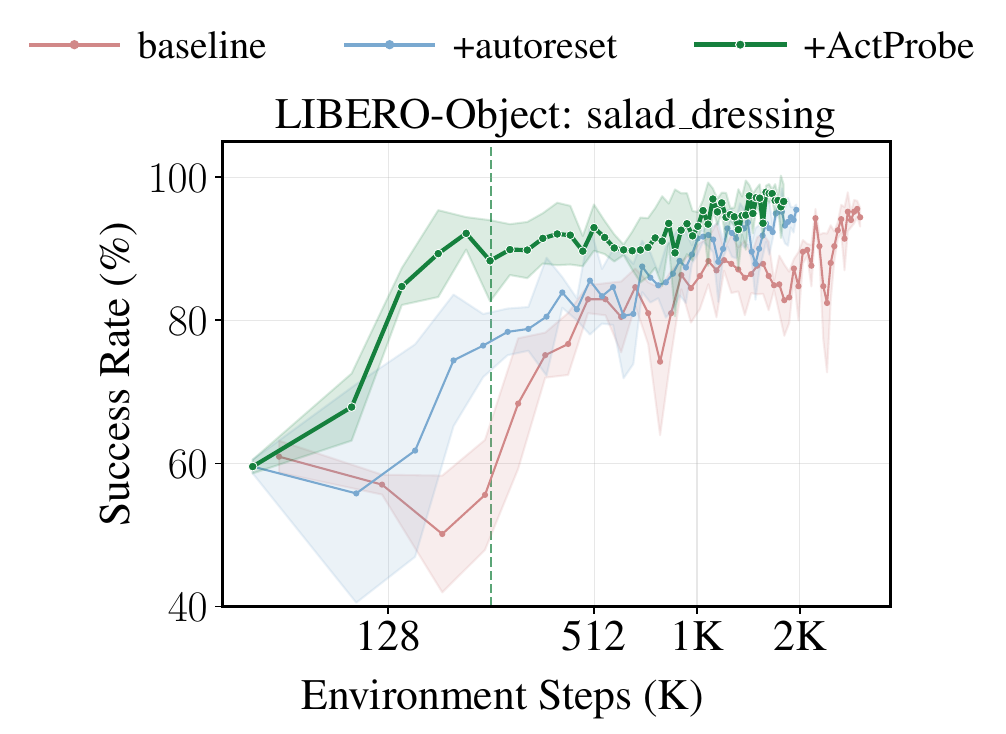}
    \caption{RL fine-tuning acceleration.}
    \label{fig:rl}
  \end{subfigure}
  \caption{\textbf{\nickname in deployment and RL fine-tuning.} \emph{(a)} Real-robot setup for the two unseen pick tasks in Table~\ref{tab:rl_robot}. \emph{(b)} Train success rate vs.\ cumulative environment steps on LIBERO-Object (\emph{salad dressing}), 3 seeds (shaded std), for default RLinf, auto-reset, and auto-reset $+$ \nickname{} early termination (Appendix~\ref{app:rl}). \nickname{} reaches baseline-level performance with $2.9\times$ fewer environment interactions.}
  \label{fig:applications}
\end{figure}

\subsection{Applications}
\label{sec:experiment_discard}

We evaluate two downstream uses of \nickname, in both cases deploying the same two-feature probe used in our offline benchmarks with no additional training: (a) zero-shot real-robot deployment (Figure~\ref{fig:robot_setup}, Table~\ref{tab:rl_robot}) and (b) acceleration of RL fine-tuning (Figure~\ref{fig:rl}).

\textbf{(a) \nickname{} transfers zero-shot from simulation to a real robot.} We deploy the GR00T+RoboCasa-trained probe alongside the same GR00T policy on a Unitree G1-D humanoid---a different embodiment from the Franka-arm simulator---for two unseen pick tasks (Table~\ref{tab:rl_robot}). Counting a trial as correct when the probe stays silent on a success and alerts on a failure, \nickname{} attains the most correct rollouts on both, outperforming STAC-Single and SAFE-MLP under real-world perception noise and execution shift.

\textbf{(b) \nickname{} accelerates RL fine-tuning at no cost to final performance.} We fine-tune GR00T with PPO~\cite{schulman2017ppo} on the salad dressing task from LIBERO-Object within RLinf~\cite{yu2025rlinf}, where \nickname{} ends high-confidence failures early and restarts fresh rollouts rather than running them to timeout. Because the reward is sparse and delivered only at the final step, truncating a predicted failure discards little learning signal, so the reclaimed interactions yield more informative rollouts---matching the baseline's final success rate with $2.9\times$ fewer environment interactions (Figure~\ref{fig:rl}).

% !TEX root = ../neurips_2026.tex
% ===========================================================================
% 7. CONCLUSION
% ===========================================================================
\section{Conclusion}
\label{sec:conclusion}

We presented \nickname{}, a lightweight action-space probe that anticipates failures in generative robot policies, showing that two compact signals---Temporal Consistency Error (TCE) and Action Chunk Magnitude (ACM)---are suffic for failure detection. The signals are mapped through a task-conditioned bridged LSTM--MLP designed for early detection, combining instantaneous and temporally integrated cues. Across five policy--environment settings, \nickname{} improves the F1--timeliness Pareto frontier by an average hypervolume gain of $+12.7\%$ over prior baselines.

\bibliographystyle{unsrtnat}
\bibliography{references}

% !TEX root = ../neurips_2026.tex
% ===========================================================================
% APPENDIX
% ===========================================================================
\appendix

% -------------------------------------------------------------------
\section{Architecture and training hyperparameters}
\label{app:training}

\paragraph{Training loss.}
The probe is trained with the per-step binary cross-entropy, averaged over valid (non-padded) timesteps:
\[
\mathcal{L} \;=\; -\frac{1}{\sum_i T_i} \sum_i \sum_{t=1}^{T_i} \bigl[\, y_i \log s_{i,t} + (1 - y_i)\log(1 - s_{i,t})\,\bigr],
\]
where $i$ indexes training episodes of length $T_i$, and $y_i \in \{0,1\}$ is the episode-level outcome label broadcast to every valid timestep of episode $i$.

\paragraph{Optimization.}
We use AdamW with learning rate $10^{-3}$ and weight decay $10^{-4}$, together with a cosine annealing schedule over 400 epochs and gradient clipping at 1.0. Training uses batch size 64 and monitors validation ROC-AUC every 5 epochs; early stopping triggers after 50 epochs without an improvement larger than $10^{-4}$. In the seen-task protocol, episodes are partitioned into stratified 70/15/15 train/val/test splits within each task and outcome class. In the unseen-task protocol, train and val episodes are drawn only from seen tasks using the same 70/15 split, while all episodes from the held-out tasks form the test set. Reported numbers average over 3 random seeds $\{0,1,2\}$.

\paragraph{Architecture details.}
The final \nickname model uses the two normalized action-space features from Section~\ref{sec:features} together with a normalized timestep $t/(T-1)$ ($T$ is the episode timeout), giving a 3-dimensional per-step input. A frozen 1024-dimensional Qwen3 task embedding is projected through $\mathrm{Linear}(1024,16)$ followed by ReLU, and the resulting 16-dimensional bottleneck is mapped through two separate $\mathrm{Linear}(16,32)$ layers, one each for the LSTM hidden and cell states. The temporal encoder is a single-layer LSTM with input size 3 and hidden size 32. Dropout with rate 0.4 is applied to the LSTM outputs. The score head is an MLP with layers $\mathrm{Linear}(35,16)$, ReLU, $\mathrm{Linear}(16,8)$, ReLU, and $\mathrm{Linear}(8,1)$, followed by a sigmoid. The full detector contains approximately 24K trainable parameters and is trained with PyTorch on a single CUDA GPU.

\paragraph{Threshold calibration.}
Given $n$ successful validation rollouts with episode-level scores $\{S_i\}_{i \in \mathrm{val,\,succ}}$, the alert threshold is the split conformal quantile
\[
\tau \;=\; q_{\lceil(n+1)(1-\alpha)\rceil/n}\!\bigl(\{S_i\}_{i \in \mathrm{val,\,succ}}\bigr)
\]
at a target false-positive rate $\alpha$ (the conformal significance level). Under exchangeability this guarantees that successful rollouts exceed $\tau$ with probability at most $\alpha$.

% -------------------------------------------------------------------
\section{Per-$\alpha$ metric breakdown}
\label{app:alpha_breakdown}
\begin{figure}[p]
  \centering
  \includegraphics[width=\ifdefined\arxivversion 0.9\linewidth\else\linewidth\fi]{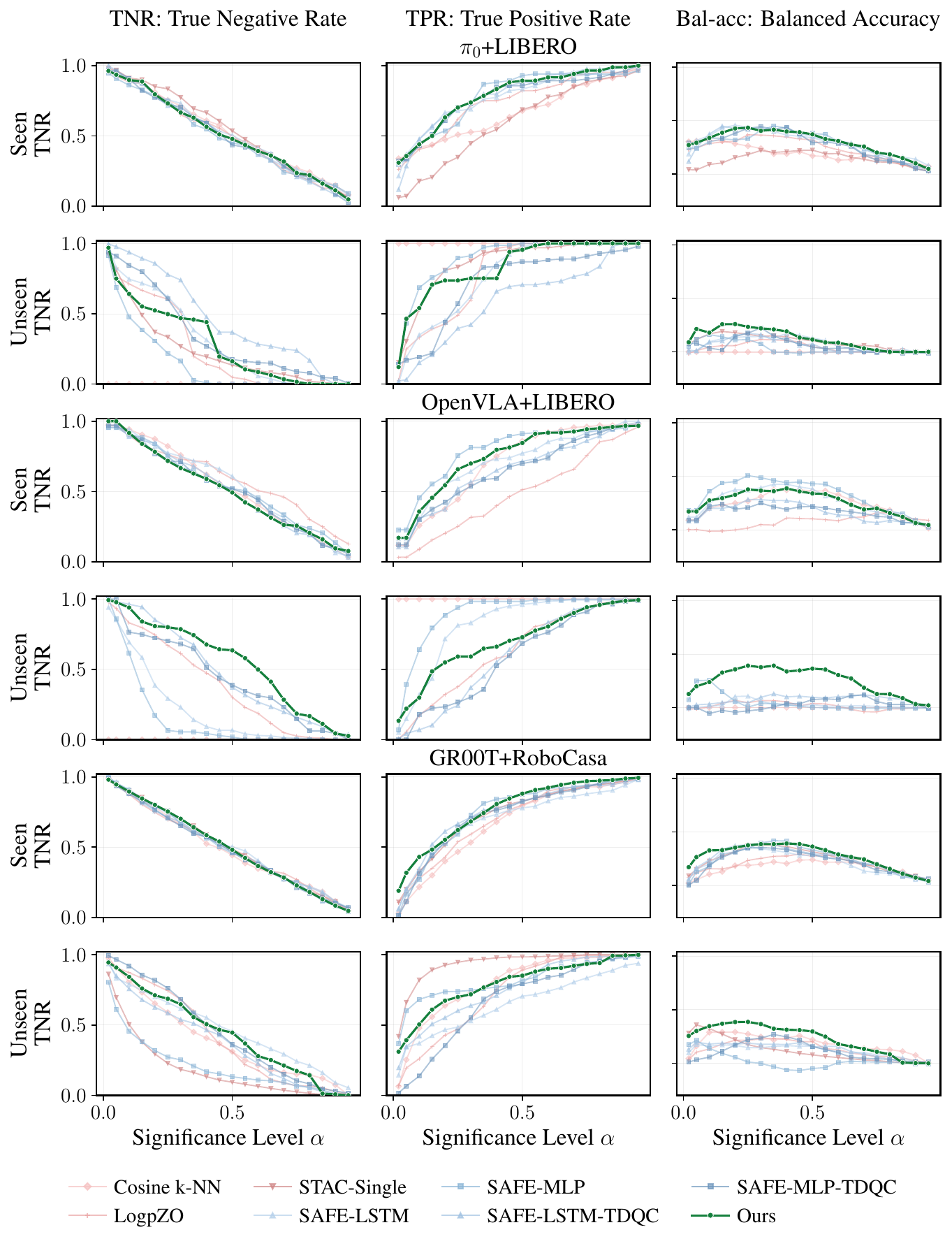}
  \caption{\textbf{Per-$\alpha$ TNR / TPR / bal-acc for $\pi_0$+LIBERO, OpenVLA+LIBERO, and GR00T+RoboCasa.} Each benchmark occupies two rows (seen, then unseen). Under the unseen protocol hidden-state baselines lose TPR sharply or collapse TNR, whereas \nickname's three curves retain shapes close to their seen-protocol counterparts.}
  \label{fig:alpha_has_unseen_p1}
\end{figure}

\begin{figure}[tbp]
  \centering
  \includegraphics[width=\linewidth]{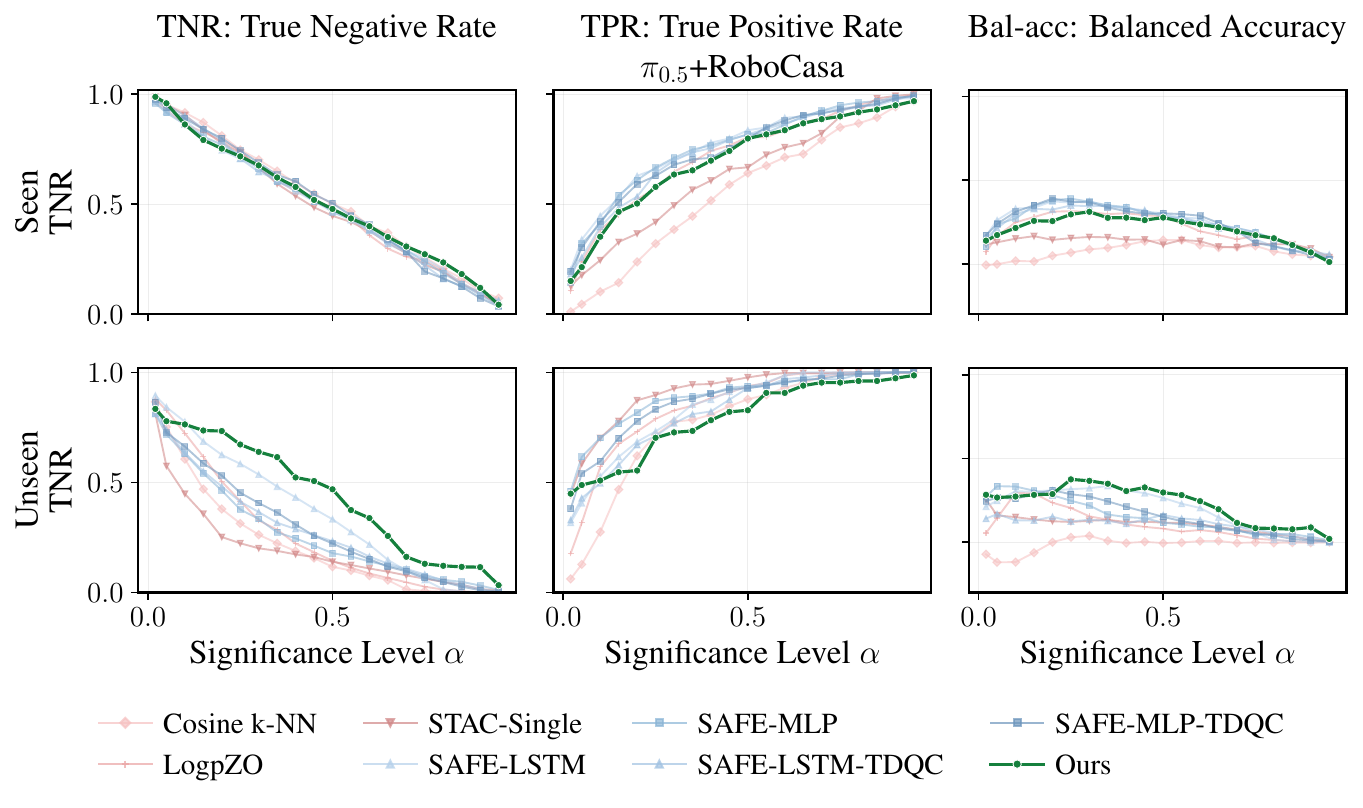}
  \caption{\textbf{Per-$\alpha$ TNR / TPR / bal-acc for $\pi_{0.5}$+RoboCasa} (continued from Figure~\ref{fig:alpha_has_unseen_p1}). The benchmark occupies two rows (seen, then unseen).}
  \label{fig:alpha_has_unseen_p2}
\end{figure}

\begin{figure}[tbp]
  \centering
  \includegraphics[width=\linewidth]{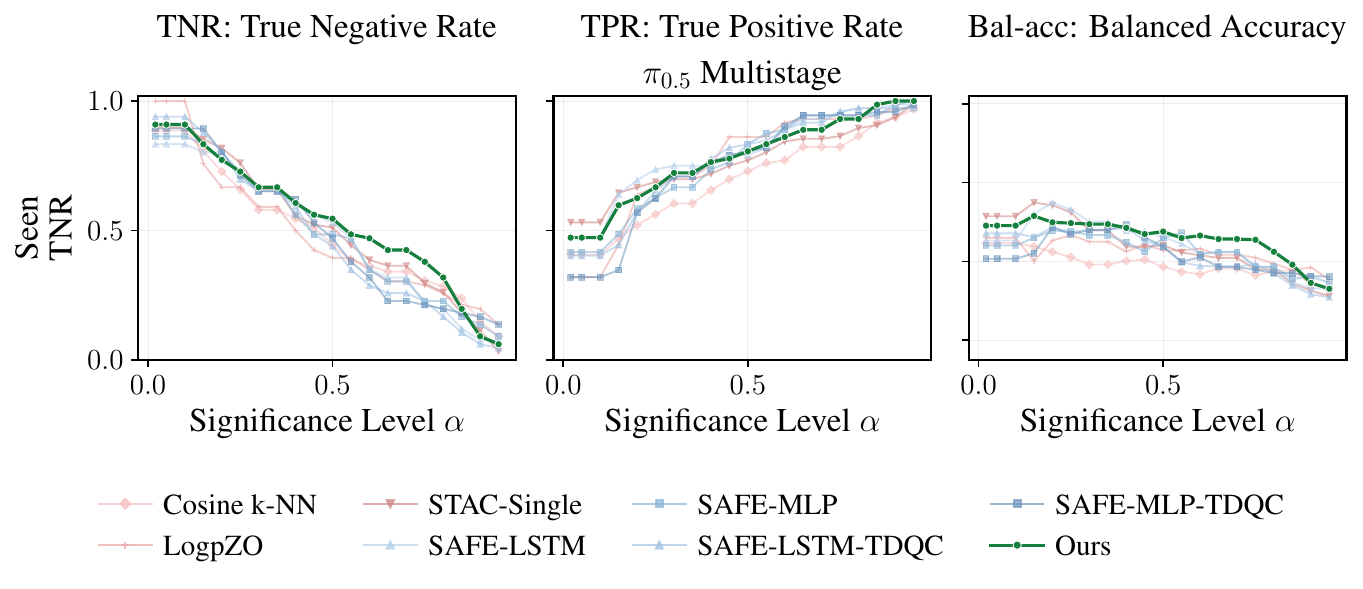}
  \caption{\textbf{Per-$\alpha$ TNR / TPR / bal-acc for the benchmark without an unseen split} ($\pi_{0.5}$ Multistage). Only the seen protocol is applicable; all six methods are reported for completeness.}
  \label{fig:alpha_no_unseen}
\end{figure}

Figures~\ref{fig:alpha_has_unseen_p1}--\ref{fig:alpha_no_unseen} decompose the false-positive-rate ($\alpha$) sweep into its three constituent metrics (TNR, TPR, balanced accuracy) for every benchmark. The four benchmarks that admit an unseen-task split are reported under both protocols, spread across two figures for legibility ($\pi_0$+LIBERO, OpenVLA+LIBERO, GR00T+RoboCasa in Figure~\ref{fig:alpha_has_unseen_p1}; $\pi_{0.5}$+RoboCasa in Figure~\ref{fig:alpha_has_unseen_p2}); $\pi_{0.5}$ Multistage has no unseen partition and appears separately under the seen protocol (Figure~\ref{fig:alpha_no_unseen}). Across all settings \nickname maintains the highest balanced accuracy at virtually every $\alpha$, and its TPR rises faster than any baseline while its TNR drops no faster than that of the hidden-state probes.

\clearpage

% -------------------------------------------------------------------
\section{Ablation studies}
\label{app:ablations}

This appendix complements the main-text ablation in Section~\ref{sec:two_feature_signal_new}. While Section~\ref{sec:two_feature_signal_new} isolates the final two-feature pipeline on $\pi_{0.5}$+RoboCasa, the ablations here provide broader supporting evidence on two additional axes: exploratory pre-redesign feature pools (Appendix~\ref{app:abl-feat}) and alternative training-label strategies (Appendix~\ref{app:abl-label}). All ablations are run on three benchmarks ($\pi_0$+LIBERO, OpenVLA+LIBERO, GR00T+RoboCasa) under both seen- and unseen-task protocols with $3$ seeds. We report three metrics per configuration: episode-level AUC (Ep AUC) and the early-detection AUCs at $q{=}0.25$ and $q{=}0.50$.

\subsection{Exploratory feature-pool ablation}
\label{app:abl-feat}

This ablation predates the final two-feature redesign and is included only to document the broader feature pools considered during development. Its main takeaway is negative but informative: adding more physical indicators does not yield stable gains across policies or protocols, and often hurts unseen-task robustness. This observation motivated the final design choice to focus on compact action-output signals.

Here, \textsc{ActProbe} denotes the final compact feature set used in the main text (TCE and ACM). \textsc{ActProbe+EEF State} augments \textsc{ActProbe} with six end-effector state variables: absolute position $(x,y,z)$ and velocity $(v_x,v_y,v_z)$. For OpenVLA, whose rollout logs do not expose real-time end-effector state, these quantities are approximated from the commanded translations and their temporal differences.

\begin{table}[h]
  \caption{\textbf{Feature group ablation, seen-task protocol.}}
  \label{tab:abl-feat-seen}
  \centering
  \small
  \setlength{\tabcolsep}{4pt}
  \begin{tabular}{llccc}
    \toprule
    Benchmark & Feature set & $q{=}0.25$ & $q{=}0.50$ & Ep AUC \\
    \midrule
    \multirow{2}{*}{$\pi_0$+LIBERO}
      & \textsc{ActProbe} & 79.5 & 84.5 & 91.6 \\
      & \textsc{ActProbe+EEF State} & 79.9 & 85.8 & 92.5 \\
    \midrule
    \multirow{2}{*}{OpenVLA+LIBERO}
      & \textsc{ActProbe} & 76.5 & 84.4 & 98.0 \\
      & \textsc{ActProbe+EEF State} & 73.5 & 83.1 & 97.4 \\
    \midrule
    \multirow{2}{*}{GR00T+RoboCasa}
      & \textsc{ActProbe} & 76.9 & 87.4 & 98.6 \\
      & \textsc{ActProbe+EEF State} & 77.2 & 88.5 & 98.9 \\
    \bottomrule
  \end{tabular}
\end{table}

\begin{table}[h]
  \caption{\textbf{Feature group ablation, unseen-task protocol.}}
  \label{tab:abl-feat-unseen}
  \centering
  \small
  \setlength{\tabcolsep}{4pt}
  \begin{tabular}{llccc}
    \toprule
    Benchmark & Feature set & $q{=}0.25$ & $q{=}0.50$ & Ep AUC \\
    \midrule
    \multirow{2}{*}{$\pi_0$+LIBERO}
      & \textsc{ActProbe} & 74.7 & 80.0 & 87.9 \\
      & \textsc{ActProbe+EEF State} & 74.7 & 80.0 & 89.8 \\
    \midrule
    \multirow{2}{*}{OpenVLA+LIBERO}
      & \textsc{ActProbe} & 73.0 & 81.2 & 99.1 \\
      & \textsc{ActProbe+EEF State} & 62.6 & 74.1 & 97.6 \\
    \midrule
    \multirow{2}{*}{GR00T+RoboCasa}
      & \textsc{ActProbe} & 73.3 & 84.7 & 98.7 \\
      & \textsc{ActProbe+EEF State} & 68.9 & 76.9 & 92.1 \\
    \bottomrule
  \end{tabular}
\end{table}

\subsection{Training label strategy}
\label{app:abl-label}

This ablation validates the uniform label-broadcast strategy used in the final method by comparing it against a ramp target that increases over time within failure episodes. We compare two failure-label strategies with all other settings held fixed: \textsc{uniform} assigns $y_t{=}1$ at every timestep of a failure episode, whereas \textsc{ramp} assigns $y_t{=}t/T$, making failure supervision weak early in the rollout and stronger near the end.

Tables~\ref{tab:abl-label-seen} and~\ref{tab:abl-label-unseen} show that ramp and uniform labels yield comparable performance across the three benchmarks, with neither strategy dominating across metrics. Ramp tends to improve Ep AUC and $q{=}0.50$, while uniform is more consistent on the earliest $q{=}0.25$ metric in seen-task settings. We retain uniform in the final method for its simplicity, while noting that ramp is a competitive alternative.

\begin{table}[h]
  \caption{\textbf{Training label strategy, seen-task protocol.}}
  \label{tab:abl-label-seen}
  \centering
  \small
  \setlength{\tabcolsep}{4pt}
  \begin{tabular}{llccc}
    \toprule
    Benchmark & Label & $q{=}0.25$ & $q{=}0.50$ & Ep AUC \\
    \midrule
    \multirow{2}{*}{$\pi_0$+LIBERO}
      & \textsc{uniform} & 79.5 & 84.5 & 91.6 \\
      & \textsc{ramp}    & 78.8 & 86.6 & 93.1 \\
    \midrule
    \multirow{2}{*}{OpenVLA+LIBERO}
      & \textsc{uniform} & 76.5 & 84.4 & 98.0 \\
      & \textsc{ramp}    & 73.5 & 88.5 & 97.9 \\
    \midrule
    \multirow{2}{*}{GR00T+RoboCasa}
      & \textsc{uniform} & 76.9 & 87.4 & 98.6 \\
      & \textsc{ramp}    & 78.2 & 91.5 & 99.9 \\
    \bottomrule
  \end{tabular}
\end{table}

\begin{table}[h]
  \caption{\textbf{Training label strategy, unseen-task protocol.}}
  \label{tab:abl-label-unseen}
  \centering
  \small
  \setlength{\tabcolsep}{4pt}
  \begin{tabular}{llccc}
    \toprule
    Benchmark & Label & $q{=}0.25$ & $q{=}0.50$ & Ep AUC \\
    \midrule
    \multirow{2}{*}{$\pi_0$+LIBERO}
      & \textsc{uniform} & 74.7 & 80.0 & 87.9 \\
      & \textsc{ramp}    & 74.7 & 82.6 & 91.1 \\
    \midrule
    \multirow{2}{*}{OpenVLA+LIBERO}
      & \textsc{uniform} & 73.0 & 81.2 & 99.1 \\
      & \textsc{ramp}    & 73.5 & 89.6 & 99.7 \\
    \midrule
    \multirow{2}{*}{GR00T+RoboCasa}
      & \textsc{uniform} & 73.3 & 84.7 & 98.7 \\
      & \textsc{ramp}    & 77.3 & 90.9 & 99.3 \\
    \bottomrule
  \end{tabular}
\end{table}

% -------------------------------------------------------------------
\section{Benchmark and implementation details}
\label{app:details}

\paragraph{Benchmark summary.}
Table~\ref{tab:app-benchmarks} summarizes the five policy--environment settings used throughout the evaluation. Both LIBERO~\cite{liu2023libero} and RoboCasa single-stage~\cite{nasiriany2024robocasa} support seen-task and unseen-task evaluation via 70/30 task splits. The RoboCasa composite setting is evaluated only under the seen-task protocol, since it uses a five-task subset spanning pantry restocking, arrange vegetables, cookware soaking, microwave thawing, and coffee preparation.

\begin{table}[h]
  \centering
  \caption{\textbf{Benchmark overview.} Task counts in the Seen/Unseen columns denote the number of task identities used under each protocol; ``---'' indicates that only the seen-task protocol is applicable.}
  \label{tab:app-benchmarks}
  \small
  \begin{tabular}{llllcc}
    \toprule
    Benchmark & Policy & Environment & Action type & Seen tasks & Unseen tasks \\
    \midrule
    LIBERO      & $\pi_0$       & LIBERO    & Chunk  & 10 & 3 \\
    LIBERO      & OpenVLA       & LIBERO    & AR     & 10 & 3 \\
    RoboCasa$_{\mathrm{S}}$ & GR00T         & RoboCasa  & Chunk  & 24 & 7 \\
    RoboCasa$_{\mathrm{S}}$ & $\pi_{0.5}$   & RoboCasa  & Chunk  & 24 & 7 \\
    RoboCasa$_{\mathrm{L}}$ & $\pi_{0.5}$   & RoboCasa  & Chunk  & 5  & --- \\
    \bottomrule
  \end{tabular}
\end{table}

\paragraph{Rollout collection.}
Each task contributes up to 50 rollout episodes. All detectors share identical episode sets; training, validation, and test splits are fixed across methods to ensure comparable evaluation. For the seen-task protocol, episodes are partitioned 70/15/15 per task; for the unseen-task protocol, the held-out tasks' episodes form the test set in their entirety.

% -------------------------------------------------------------------
\section{Baseline implementation details}
\label{app:baselines}

\paragraph{White-box (hidden-state) baselines.}
The SAFE family~\cite{gu2025safe} uses per-step hidden-state embeddings extracted from the policy's transformer backbone. SAFE-MLP applies a per-step MLP classifier on the embedding sequence, while SAFE-LSTM aggregates the same embeddings through a single-layer LSTM before classification. The TDQC variants~\cite{francis2026temporal} retain identical input modality and probe backbone but replace standard rollout-success supervision with temporal-difference calibration targets. We use the released SAFE codebase for both feature extraction and classifier training, with default hyperparameters from the original authors. Cosine $k$-NN serves as a non-parametric hidden-state baseline: at inference time, the per-step embedding is compared against a memory bank of training embeddings via cosine similarity, and the resulting score is calibrated under the same conformal protocol as the other detectors.

\paragraph{Black-box action-space baselines.}
LogpZO~\cite{xu2025logpzo} follows its original score-based failure-detection formulation, fitting a per-task model over action sequences and triggering when the running log-probability falls outside the calibration distribution. STAC-Single~\cite{agia2024stac} is the strongest real-time variant of STAC reproduced in the SAFE codebase: it uses the policy's native single-sample action chunks (no repeated resampling, no VLM progress module) and computes disagreement on the overlapping horizon between consecutive chunks. FIPER~\cite{romer2025fiper} is included in the architecture ablation (Section~\ref{sec:two_feature_signal_new}) as an action-resampling baseline; because it requires repeated forward passes per inference step, we report only its $q{=}0.25$ AUC.

% -------------------------------------------------------------------
\section{Real-robot experiment details}
\label{app:robot}

\paragraph{Hardware setup.}
The real-robot study performs zero-shot deployment of the GR00T policy trained on RoboCasa onto a Unitree G1-D dual-arm humanoid robot for tabletop manipulation (Figure~\ref{fig:robot_setup}). The robot takes RGB observations from onboard cameras and executes the policy in closed loop at the nominal control frequency. During data collection, only the policy is run online; action chunks and trial outcomes are logged for subsequent detector evaluation.

\paragraph{Tasks.}
We evaluate two unseen pick tasks in a shared tabletop workspace. In \emph{Pick Yellow Pear}, the robot must grasp a yellow pear and place it into a small dish. In \emph{Pick Milk Carton}, the target object is replaced by a milk carton while the rest of the setup is unchanged. Neither target object nor the dish appears in the RoboCasa training rollouts used for policy or detector development.

\paragraph{Trial protocol.}
Each task is evaluated for 12 independent trials. Object poses are re-randomized within a fixed tabletop region between trials, while the dish remains fixed. A trial is counted as physically successful if the target object is grasped and stably placed in the dish before the nominal episode budget expires. After online collection, each detector is run offline on the recorded action sequence to determine whether its running-maximum score would have crossed the calibrated threshold. In the detector-level comparison of Table~\ref{tab:rl_robot}, a trial counts as correct for a detector if it stays below threshold on a physically successful rollout, or if it raises an alert during a physically failed rollout.

\paragraph{Probe deployment.}
The deployed \nickname probe is exactly the same two-feature configuration used in the offline benchmarks. Its conformal threshold is calibrated entirely on offline RoboCasa rollouts at $\alpha{=}0.15$, with no real-robot data used for training or threshold selection. SAFE-MLP and STAC-Single are applied in the same offline replay protocol using the same training source and calibration procedure, so the comparison isolates detector design rather than data-access differences.

% -------------------------------------------------------------------
\section{RL fine-tuning details}
\label{app:rl}

\paragraph{Setup.}
We fine-tune the GR00T policy with PPO~\cite{schulman2017ppo} on the \emph{salad dressing} task from LIBERO-Object within the RLinf framework~\cite{yu2025rlinf}, where the reward is sparse---a single terminal reward on task completion, with no per-step shaping. We report train success rate against cumulative environment steps, averaged over 3 seeds; full PPO and \nickname hyperparameters are in Table~\ref{tab:rl_hparams}.

\paragraph{Rollout schedules.}
We compare three rollout schedules under a matched total interaction budget (the three curves in Figure~\ref{fig:rl}). The budget counts every environment step actually executed (summed across parallel environments), including steps in rollouts later truncated by \nickname; the schedules differ only in how they \emph{spend} this fixed budget. (a) \emph{Default RLinf}: an episode ends on success or timeout, but an environment that finishes early (on success) idles until the fixed rollout horizon before a new episode begins, leaving an inter-rollout bubble. (b) \emph{Auto-reset} removes inter-rollout bubbles by letting an environment start a fresh rollout immediately once the current episode succeeds, without waiting out the remaining timeout. (c) \emph{Auto-reset $+$ \nickname} additionally ends rollouts that \nickname flags as high-confidence failures early, reallocating the reclaimed interactions to new rollouts. We report the $2.9\times$ factor as the ratio of cumulative environment interactions at which schedule (c) first reaches the final (plateau) success rate of the default schedule (a).

\begin{table}[!t]
\centering
\caption{\textbf{Hyperparameters for RL fine-tuning} (GR00T~$+$~PPO on LIBERO-Object \emph{salad dressing}). Left: PPO and policy optimization. Right: the \nickname online detector used by the \emph{auto-reset $+$ \nickname} schedule.}
\label{tab:rl_hparams}
\small
\begin{tabular}{@{}ll@{\hspace{2em}}ll@{}}
\toprule
\multicolumn{2}{l}{\textbf{PPO / policy}} & \multicolumn{2}{l}{\textbf{\nickname online detector}} \\
\midrule
Policy backbone        & GR00T ($+$ value head) & Retrain interval        & every 1 PPO step \\
Action-chunk size      & 5 env steps            & Retrain epochs          & 50 \\
Episode timeout        & 240 steps (48 chunks)  & Online buffer capacity  & 400 episodes \\
Parallel environments  & 32                     & Base (frozen) episodes  & 200 SFT rollouts \\
Optimizer              & AdamW                  & Min.\ episodes to start & 50 \\
Policy / value LR      & $2{\times}10^{-5}$     & Warmup (flag-only)      & 2 steps \\
Adam $(\beta_1,\beta_2)$ & $(0.9, 0.95)$        & Detection window $K$    & 1 \\
Weight decay           & $0.01$                 & Window aggregation      & median \\
Gradient clip          & $1.0$                  & Min.\ cut chunk         & 10 \\
Discount $\gamma$      & $0.99$                 & Spare (cut) ratio       & $0.7$ \\
GAE $\lambda$          & $0.95$                 & Threshold $\tau$ init   & $0.95$ \\
PPO clip $\epsilon$    & $0.2$                  & $\tau$ recalibration    & $P_{95}^{\mathrm{succ}}{+}0.02$ \\
Value clip             & $0.2$                  & Language embedding      & Qwen3 (1024-d) \\
Advantage              & GAE, normalized        & Detector params         & ${\approx}24$K \\
PPO epochs / update    & 1                      & Retrain placement       & background thread \\
Global / micro batch   & $256 / 32$             & & \\
Entropy / KL coef      & $0 / 0$                & & \\
Sampling temp.\ (tr/ev)& $1.0 / 0.6$            & & \\
Reward                 & sparse terminal        & & \\
Seeds                  & 3                      & & \\
\bottomrule
\end{tabular}
\end{table}

\paragraph{Online co-training of the detector.}
The \nickname detector used during RL is retrained from scratch at every PPO step so it tracks the improving policy, rather than reusing the frozen offline probe. Each step's completed rollouts (successes and failures) enter a sliding buffer of recent on-policy episodes, unioned with a fixed base set of 200 SFT-policy rollouts; the base set guarantees enough complete \emph{failure} trajectories to fit a boundary even once the buffer becomes success-dominated. Retraining runs in a background thread so the PPO loop is never blocked, and the refreshed weights and threshold are swapped in at the next step (the first two steps are warmup---scoring but no cuts). The threshold is recalibrated online to the $95$th percentile of current success scores plus a small margin---the same upper-quantile rule as the offline conformal threshold (Appendix~\ref{app:training}), re-estimated each step so the operating point does not drift as success rate climbs.

\end{document}